\documentclass[11pt,a4paper]{article}
\usepackage{acl}
\usepackage{times}
\usepackage{latexsym}
\usepackage{comment}
\usepackage{amsfonts}
\usepackage{amsmath}
\usepackage{graphicx}
\usepackage{multirow}
\usepackage{booktabs}

\newcommand{\CLS}{\texttt{CLS}\ }
\newcommand{\Tuned}{\textsc{Tuned}\ }
\newcommand{\Untuned}{\textsc{Untuned}\ }
\newcommand{\TunedOrig}{\textsc{TunedOrig}\ }

\usepackage{microtype}

\title{
    Fine-Tuned Transformers Show Clusters of \\Similar Representations Across Layers
}

\author{
  Jason Phang$^1$,
  Haokun Liu$^2$,
  Samuel R. Bowman${}^1{}^3{}^4$\\\AND
\textnormal{$^1$Center for Data Science, New York University} \\
\textnormal{$^2$Dept. of Computer Science, University of North Carolina at Chapel Hill}
\\
\textnormal{$^3$Dept. of Linguistics, New York University}
\\
\textnormal{$^4$Dept. of Computer Science, New York University}
\AND Correspondence: \href{mailto:jasonphang@nyu.edu}{\tt jasonphang@nyu.edu}
}

\date{}

\begin{document}
\maketitle
\begin{abstract}


Despite the success of fine-tuning pretrained language encoders like BERT for downstream natural language understanding (NLU) tasks, it is still poorly understood how neural networks change after fine-tuning.
In this work, we use centered kernel alignment (CKA), a method for comparing learned representations, to measure the similarity of representations in task-tuned models across layers.
In experiments across twelve NLU tasks, we discover a consistent block diagonal structure in the similarity of representations within fine-tuned RoBERTa and ALBERT models, with strong similarity within clusters of earlier and later layers, but not between them.
The similarity of later layer representations implies that later layers only marginally contribute to task performance, and we verify in experiments that the top few layers of fine-tuned Transformers can be discarded without hurting performance, even with no further tuning.


\end{abstract}

\section{Introduction}

Fine-tuning pretrained language encoders such as BERT \citep{devlin-etal-2019-bert} and its successors \citep{liu2019roberta, Lan2020ALBERT, clark2020electra, he2020deberta} has proven to be highly successful, attaining state-of-the-art performance on many language tasks, but how do these models internally represent task-specific knowledge?

In this work, we study how learned representations change through fine-tuning by studying the similarity of representations between layers of untuned and task-tuned models. We use centered kernel alignment \citep[CKA;][]{kornblith2019cka} to measure representation similarity and conduct extensive experiments across three pretrained encoders and twelve language understanding tasks.

\begin{figure}[t]
    \centering
    \includegraphics[width=0.45\textwidth]{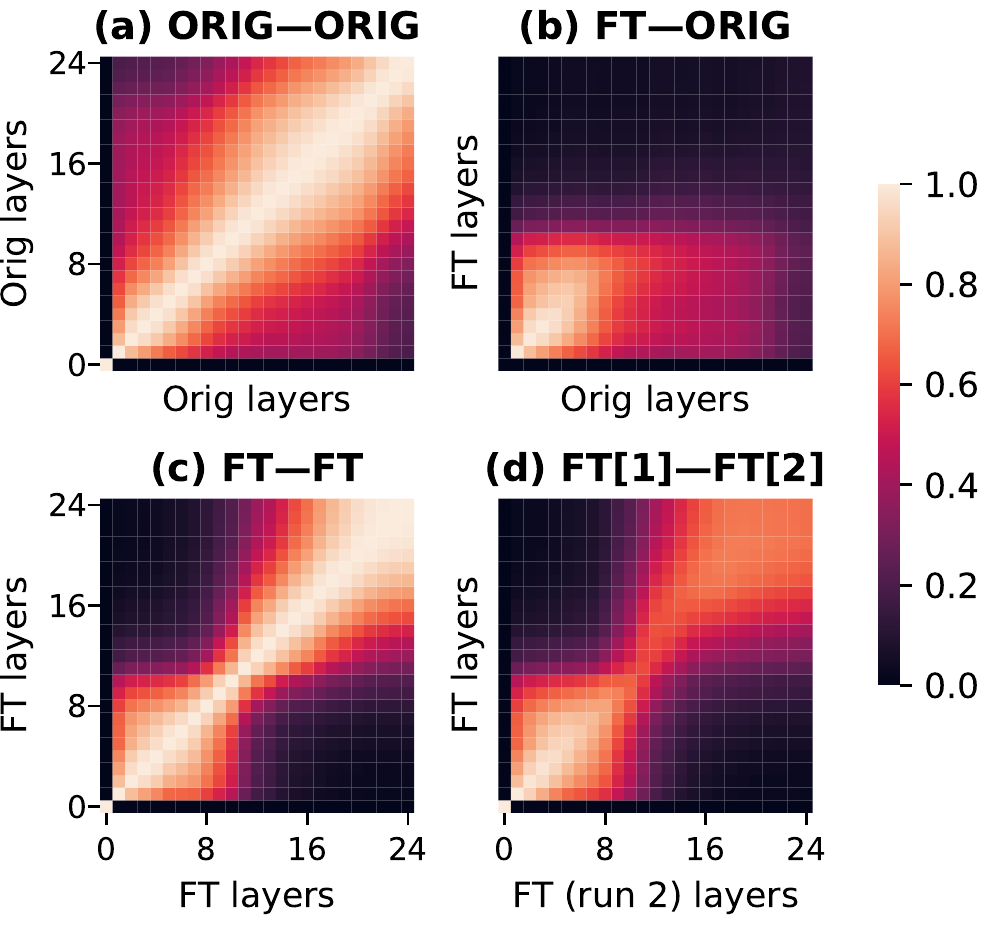}
    \caption{CKA similarity scores of \CLS (classifier token) representations of ORIG (untuned ALBERT) and FT (fine-tuned) models on RTE, across different layers of the model. FT[1]--FT[2] compares two RTE models with different random restarts. ORIG--ORIG and FT--FT are symmetric by construction. Fine-tuned models exhibit a block-diagonal structure in the representation similarities. The same color scale is used in all plots.}
    \label{fig:cka-example}
\end{figure}

We discover a consistent, block diagonal structure (Figure~\ref{fig:cka-example}c,d) in the similarity of learned representations for almost all task-tuned RoBERTa and ALBERT models, where early layer representations and later layer representations form two distinct clusters, with high intra-cluster and low inter-cluster similarity.

Given the strong representation similarity of later model layers, we hypothesize that many of the later layers only marginally contribute to task performance. We show in experiments that the later layers of task-tuned RoBERTa and ALBERT can indeed be discarded with minimal impact to performance, even without any further fine-tuning.

\section{Experimental Setup}
\label{experimental_setup}

\paragraph{Models}
For the majority of our experiments, we consider three commonly used language-encoding models: RoBERTa \citep{liu2019roberta}, ALBERT \citep{Lan2020ALBERT} and ELECTRA \citep{clark2020electra}. Because of the large number of experiments being performed, we use RoBERTa\textsubscript{\textsc{Base}}, ALBERT\textsubscript{\textsc{LargeV2}} and ELECTRA\textsubscript{\textsc{Base}} rather than the largest available versions of these models.

\paragraph{Tasks}
We use the tasks included in the GLUE benchmark \citep{wang-etal-2018-glue} excluding the data-poor WNLI,
namely: CoLA \citep{warstadt2018cola}, MNLI \citep{williams2018mnli}, MRPC \citep{dolan2005mrpc}, QNLI \cite{rajpurkur2016squad}, QQP,\footnote{\url{https://quoradata.quora.com/First-Quora-Dataset-Release-Question-Pairs} \label{qqp-source}} RTE \citep{dagan2005rte}, SST-2 \citep{socher2013sst2}, and STS-B \citep{cer2017stsb}. We include four additional tasks to cover a more diverse set of task formats and difficulties: BoolQ \citep{clark-etal-2019-boolq} and Yelp Review Polarity \citep{zhang2015yelp} classification tasks, and HellaSwag \citep{zellers2019hellaswag} and CosmosQA \citep{huang2019cosmosqa} multiple-choice tasks.

\paragraph{Optimization}
The representations learned over the course of training and similarity of representations may be sensitive to the number of steps used in training. To control for this, and to avoid task-specific hyperparameter tuning, we fine-tune on each task for up to 10,000 steps. We use the Adam \citep{kingma2014adam} optimizer with batch size of 4, a learning rate of 1e-5, and 1,000 warmup optimization steps.

We use the \texttt{jiant} \citep{phang2020jiant} library, built on Transformers \citep{wolf-etal-2020-transformers} and PyTorch \citep{pytorch2019}, to run our experiments. 

\section{Representation Similarity with CKA}

To analyze how learned representations change via fine-tuning, we use centered kernel alignment \citep[CKA;][]{kornblith2019cka} to measure representation similarity. CKA is invariant to both orthogonal transformation and isotropic scaling of the compared representations, making it ideal for measuring the similarity of neural network representations, and has applied to BERT-type models in prior work \citep{wu2020similarity,sridhar2020undivided}. Given two sets of representations $X \in \mathbb{R}^{N\times d_1}$ and $Y \in \mathbb{R}^{N\times d_1}$ where $N$ is the number of examples and $d_1,d_2$ the hidden dimensions, CKA computes a similarity score between 0 and 1, where a higher score indicates greater similarity. Further details on CKA are provided in Appendix~\ref{appendix:cka}.

Using CKA, we can compare the similarity of representations between different layers of the same model or even different models. For our analysis, we use the representations of the \CLS token, i.e. the token whose final layer representation is fed to the task output head.\footnote{RoBERTa uses a \texttt{<s>} token instead, 
but for brevity and consistency, we will refer to it as \CLS as well.
} We compute CKA over the validation examples of each task.


\label{section:visualizing}

\begin{figure*}[t]
    \centering
    \includegraphics[width=1.0\textwidth]{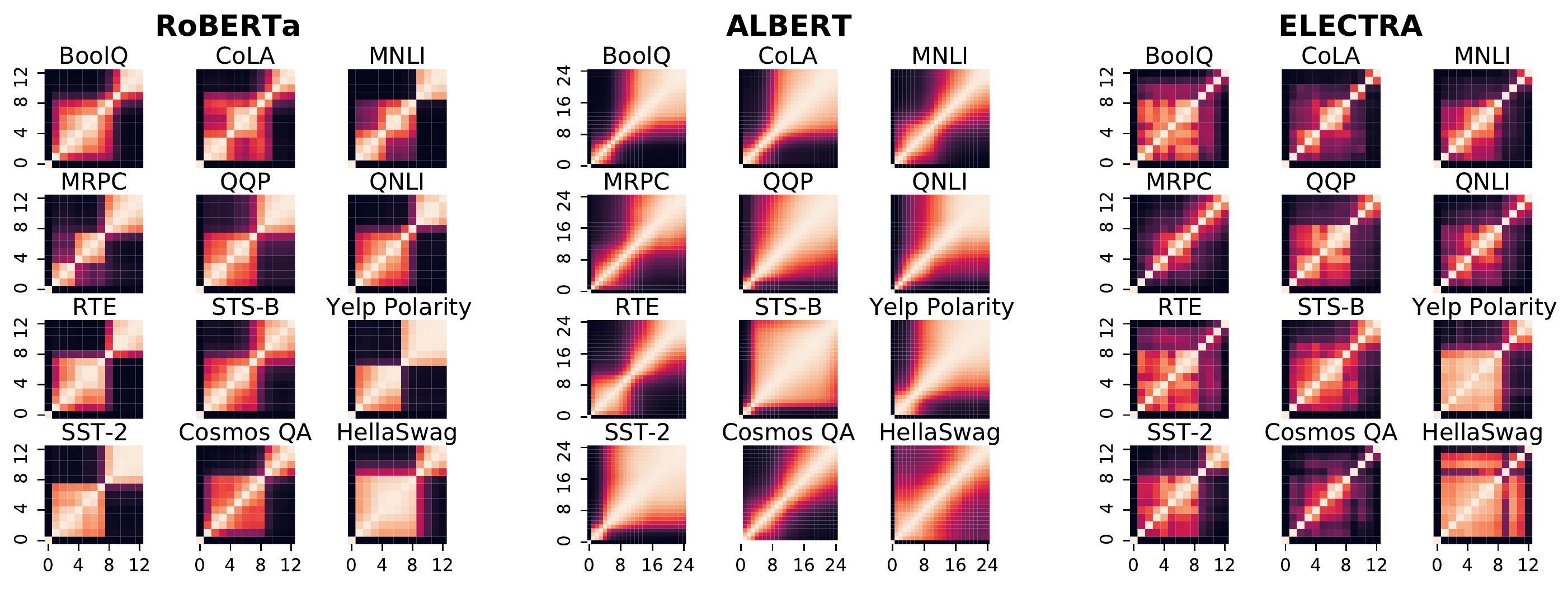}
    \caption{
        Representation similarity between layers for task-tuned models (FT--FT).
        RoBERTa and ALBERT task models exhibit a `block diagonal` structure in the representation similarity of \CLS tokens across nearly all tasks.
    }
    \label{fig:cka-tasktask-cls}
\end{figure*}

To provide intuition for CKA scores, we first show in Figure~\ref{fig:cka-example} an example of the comparison formats using ALBERT fine-tuned on RTE.

\paragraph{ORIG--ORIG} The top left plot shows the similarity of representations across the layers of the untuned ALBERT model on RTE inputs. Adjacent layers have high similarity scores, only gradually decreasing as more distant layers are compared.

\paragraph{FT--ORIG} We show layers of the task-tuned model on the Y-axis and untuned model on the X-axis. The \CLS representations of the later layers in the task-tuned model appear highly dissimilar to any of the untuned model: In other words, the representations differ starkly from those used for ALBERT's masked language modeling (MLM) and sentence order prediction (SOP) pretraining. This coheres with prior work showing that representations of later layers are most likely to change during fine-tuning \citep{kovaleva-etal-2019-revealing, wu2020similarity}.

\paragraph{FT--FT} Next, we compare layers within a single fine-tuned model. 
We observe a block-diagonal structure in the representation similarities---two distinct clusters of earlier (approx. first 10) and later (approx. last 14) layers 
that have high inter-cluster but low intra-cluster similarity.
When considered together with FT-ORIG, we can infer that the earlier layer representations resemble those used for pretraining, whereas the later layers encode a representation suitable for tackling the task.
The high internal similarity between the top few layers and the sharp block diagonal structure of the similarity matrix imply that the representations starkly differ. 

\paragraph{FT[1]--FT[2]} Finally, we compare fine-tuned ALBERT models across two random restarts. We observe a similar block diagonal structure. In particular, the similarity of the \CLS representations in the later layers indicates that CKA is able recover the similarity of representations for tackling the same task across random restarts. This likely arises as the models are fine-tuned from the same initial pretrained parameters.

\subsection{Results}

We extend our CKA analysis to all twelve tasks and all three pretrained models, showing the FT-FT results in Figure~\ref{fig:cka-tasktask-cls}. We observe that the block diagonal structure of representation similarity identified in Section~\ref{section:visualizing} appears in almost every RoBERTa and ALBERT model, sharply delineating the earlier and later clusters of representations. In fact, RoBERTa often has even more distinct clusters than ALBERT. We hypothesize that since ALBERT shares parameters across layers, it is more difficult for representations to sharply change across a single layer, whereas RoBERTa, which has no parameter sharing, has no such constraint.

The significant similarity of the later layers suggests that many of the later layers may not contribute much to the task. 
Given residual connections between Transformer layers, later layers could learn a `no-op' or only slightly adjust the output representation if the task can be adequately `solved' at an earlier layer.
If this is true, we should be able to feed an intermediate representation from later layers to the output head with no further fine-tuning and retain most of the task performance. We investigate this hypothesis in Section~\ref{section:highway}.


In contrast, we do not see the same pattern in the ELECTRA models. 
The representations of the later layers are generally highly dissimilar even up to the penultimate layer in many tasks. 
A few tasks do exhibit a minor block diagonal structure, such as STS-B, Yelp Polarity and SST-2, but it is far less apparent compared to the other two models.
ELECTRA has a very different pretraining task from the other two models (replaced token detection), which may explain this difference.

We see complementary results for FT--ORIG and FT[1]--FT[2] in Figure~\ref{fig:cka-taskbase-cls} and Figure~\ref{fig:cka-tasktask2-cls}. 
For RoBERTa and ALBERT, while the earlier layers of the task models have similar \CLS representations to the untuned models, the later layers are largely dissimilar to any layer in the base model. 

\section{Truncating Fine-tuned Models}

\label{section:highway}

To test our hypothesis that the later layers of tuned task-models only marginally contribute to task performance, we propose a simple experiment where we feed the representations from an intermediate layer directly to the task output head, effectively discarding the later layers. We refer to these as \textit{truncated} models. We test three different configurations: (a) \Untuned, where we feed intermediate representations from a fine-tuned model to the tuned task output head \textit{without any further fine-tuning}, (b) \Tuned, where we fine-tune only the output head, and (c) \TunedOrig, where we use representations from the base model (not fine-tuned on the task), but we fine-tune the output head. 
Performance of the \Untuned trunated models indicates the extent to which an intermediate representation can be directly substituted for the final layer's representation; the \Tuned and \TunedOrig models provide an upper-bound of performance using the \CLS representation of a given layer of a fine-tuned and non-fine-tuned encoder respectively.

\begin{figure*}[t]
    \centering
    \includegraphics[width=1.0\textwidth]{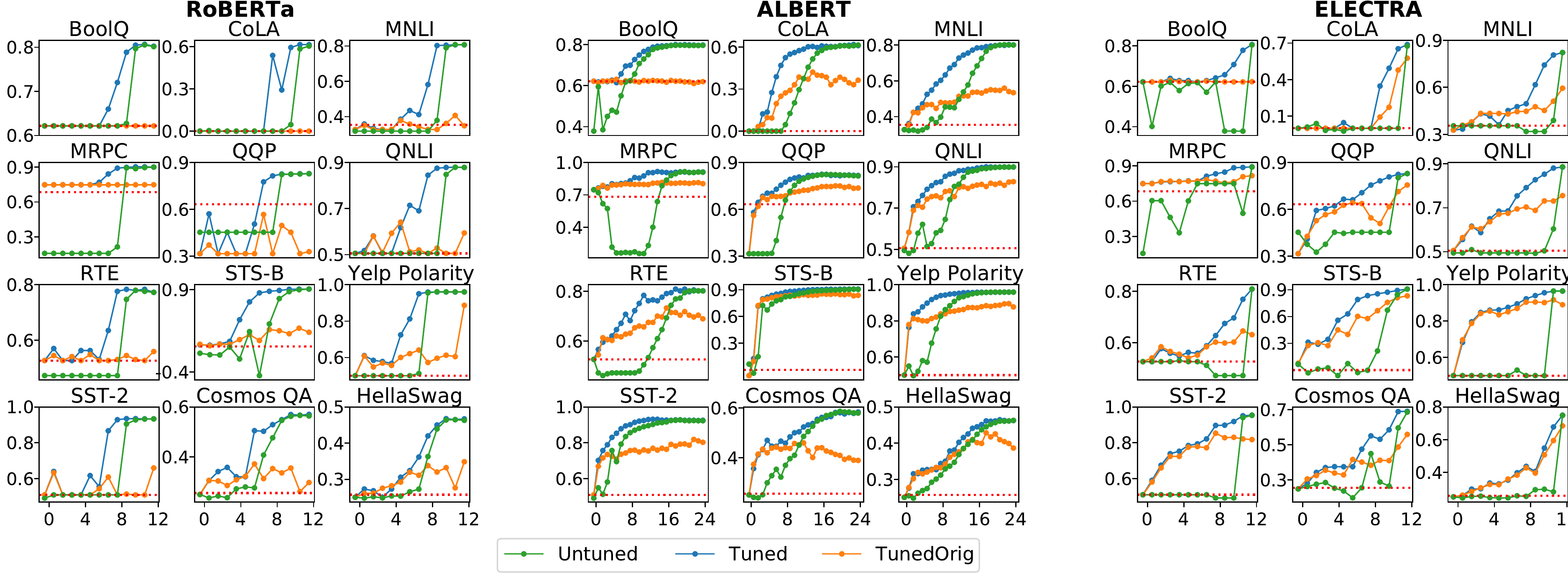}
    \caption{
        Model Truncation Experiments: Task performance (Y-axis) when feeding representation from an intermediate layer (X-axis) directly to the task output head, equivalent to discarding the top layers of the model. 
        \Untuned (green), uses a task-tuned encoder, but no further fine-tuning of the task-tuned output head. 
        \Tuned (blue), involves further fine-tuning the output head on the intermediate representation. 
        \TunedOrig (yellow) uses the pretrained encoder, but the output head is fine-tuned. 
        For RoBERTa and ALBERT, the top few layers bcan e discarded for many tasks in either \Tuned or \Untuned configurations without hurting performance.
        The majority class baseline is shown with a red dotted line, while the rightmost data-point corresponds to a full model with no truncation.
    }
    \label{fig:truncation}
\end{figure*}

\begin{figure*}[p]
    \centering
    \includegraphics[width=0.9\textwidth]{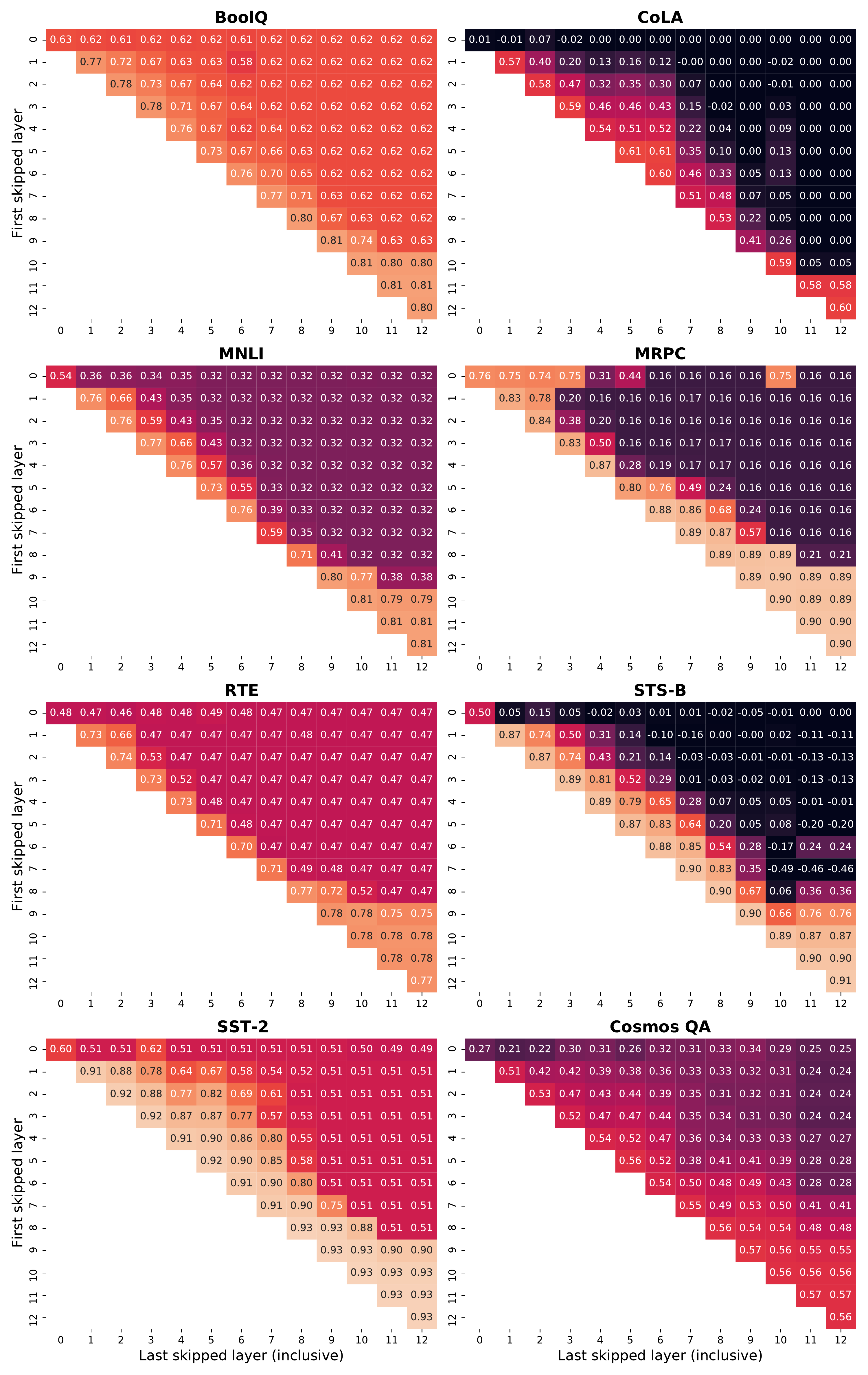}
    \caption{
        Layer Experiments: 
        Task performance when skipping contiguous spans of Transformer layers, with the Y-axis and X-axis indicating the first and last (inclusive) skipped layers, with no further fine-tuning.
        Performance tends to drop as more layers are skipped, but in many cases skipping any single layer makes little to no impact to performance, except for the first layer.
        Consistent with results above, many of the higher layers can be skipped with minimal impact to performance.
    }
    \label{fig:skip}
\end{figure*}

Our results are shown in Figure~\ref{fig:truncation}. For RoBERTa and ALBERT, we find that the \Untuned truncated models perform comparably to the Tuned truncated and full fine-tuned models\footnote{An \Untuned model using the final layer representation is equivalent to a regular fine-tuned model.} at the later layers. For instance, the top 4 layers of the RoBERTa for Yelp Polarity model can be discarded with no further tuning and minimal impact to performance (95.5 vs 96.1). 
On the other hand, \TunedOrig models perform very poorly compared to the \Tuned models across all layers, showing that task-tuned intermediate representations are crucial for good performance, even when fine-tuning the output head. 
For ALBERT, which shares parameters between layers, a larger fraction of layers can be discarded with minimal impact to performance for both \Untuned and \Tuned truncated models.

On the other hand, we do not find a similar pattern in ELECTRA models. The \Untuned truncated models perform extremely poorly when discarding almost any number of layers, and even the \Tuned truncated models quickly drop in performance with even one or two layers discarded.
These results are consistent with our CKA analyses that showed that the learned and task-tuned representations for ELECTRA do not share the same structure as those of RoBERTa and ALBERT.
We speculate that this differences stems from the different pretraining objectives---replaced token detection is a binary prediction problem, whereas masked language modeling involves predicting a distribution over a large number of tokens---leading to differences in learned representations that propagate even to fine-tuned models. We leave further investigation these differences to future work.

\subsection{Skipping Layers}

We perform a smaller set of experiments on skipping intermediate layers in a model and measuring the impact on performance.
We use fully fine-tuned RoBERTa models on a subset of the tasks we considered above, and evaluate task performance of the tuned models when we skip over contiguous spans of layers in the model without any further fine-tuning.
We show the results for skipping every possible span of layers in Fig~\ref{fig:skip}.
Performance tends to drop as larger spans of layers are skipped, although in many cases skipping any single layer seems to make little to no impact to performance.
The primary exception to this is the very first layer, where we observe that skipping just the first layer can heavily impact task performance, such as in CoLA, STS-B and Cosmos QA.
On the other hand, we find that skipping multiple of the later layers can have minimal impact on performance, consistent with our results above.
The profile of performance drops given the number of intermediate layers skipped also differs greatly across tasks: For instance, dropping more than two contiguous layers in the middle of the model seems to heavily impact MNLI and RTE performance, whereas for SST-2 the impact is not as large until 3-4 layers are skipped.

\section{Related Work}

While CKA \citep{kornblith2019cka} was initially proposed as an interpretability method for computer vision models, it has more recently seen application to NLP models. 
\citet{wu2020similarity} applied CKA to pretrained Transformers models such as BERT and GPT-2, focusing on cross-model comparison---our analysis builds on their findings, with greater focus on layer-wise comparisons and implications for fine-tuning and discarding layers.
\citet{sridhar2020undivided} use CKA to measure the impact of a proposed model architecture change on the learned representations.
\citet{voita-etal-2019-bottom} and \citet{merchant-etal-2020-happens} apply similar representation similarity analyses to Transformers, with the latter also investigating freezing and dropping layers from models.

More broadly, significant work has been done on better understanding and interpreting the capabilities of BERT-type models---\citet{rogers-etal-2020-primer} offers a thorough survey of this line of work.
Of particular relevance to our work: Work on model probing \citep{tenney2018what,liu-etal-2019-linguistic,tenney-etal-2019-bert} has studied the extent to syntactic and semantic features are represented at different layers of BERT-type models. 

Our results on model truncation also cohere with existing work on early exit in BERT models\citep{xin-etal-2020-early, xin-etal-2020-deebert, zhou2020earlyexit}, wherein models are explicitly fine-tuned to dynamically skip the later layers of a BERT encoder and directly to the output head, often to reduce inference times of models. Our results somewhat differ as we show that models can also be truncated or exited early without any explicit tuning. It has also been shown in the computer vision domain that models with residual networks work akin to an ensemble of deep and shallow models \citep{veit2016residual}.

\section{Conclusion}

We show a consistent pattern to the structure of representation similarity in task-tuned RoBERTa and ALBERT models, with strong representation similarity within clusters of earlier and later layers, but not between them. 
We further show that the later layers of task-tuned RoBERTa and ALBERT models can often be discarded without hurting task performance, verifying that the later layers of these models truly have similar representations.
However, we find that ELECTRA models exhibit starkly different properties from the other two models, which prompts further investigation into how and why these models differ.

\section*{Acknowledgments}
We would like to thank Kyunghyun Cho for his invaluable feedback on this project.
This project has benefited from financial support to SB by Eric and Wendy Schmidt (made by recommendation of the Schmidt Futures program), Samsung Research (under the project \textit{Improving Deep Learning using Latent Structure}), Apple, and Intuit, and from in-kind support by the NYU High-Performance Computing Center and by NVIDIA Corporation (with the donation of a Titan V GPU). 
This material is based upon work supported by the National Science Foundation under Grant Nos. 1922658 and 2046556. 
Any opinions, findings, and conclusions or recommendations expressed in this material are those of the author(s) and do not necessarily reflect the views of the National Science Foundation.

\bibliographystyle{acl_natbib}
\bibliography{anthology,acl2021}
\clearpage
\appendix

\section{Centered Kernel Alignment}

\label{appendix:cka}

Given two sets of representations $X \in \mathbb{R}^{N\times d}$ and $Y \in \mathbb{R}^{N\times d}$ where $N$ is the number of examples and $d$ the hidden dimension (for instance the \CLS vector representations of a set of examples from two different layers of the same model), CKA computes a similarity score between 0 and 1. :

$$ \text{CKA}(K, L) = \frac{\text{HSIC}(K, L)}{\sqrt{\text{HSIC}(K, K) \text{HSIC}(L, L) }} $$

with

$$ \text{HSIC}(K, L) = \frac{1}{(n-1)^2}\text{tr}(KHLH) $$

and $H=I_n-\frac{1}{b} \mathbf{11}^T$ $K=XX^T$, $L=YY^T$ when using a linear kernel. We refer the reader to the original work \citep{kornblith2019cka} for more details and properties of CKA.

\section{Additional Results}

\begin{figure*}[t]
    \centering
    \includegraphics[width=\textwidth]{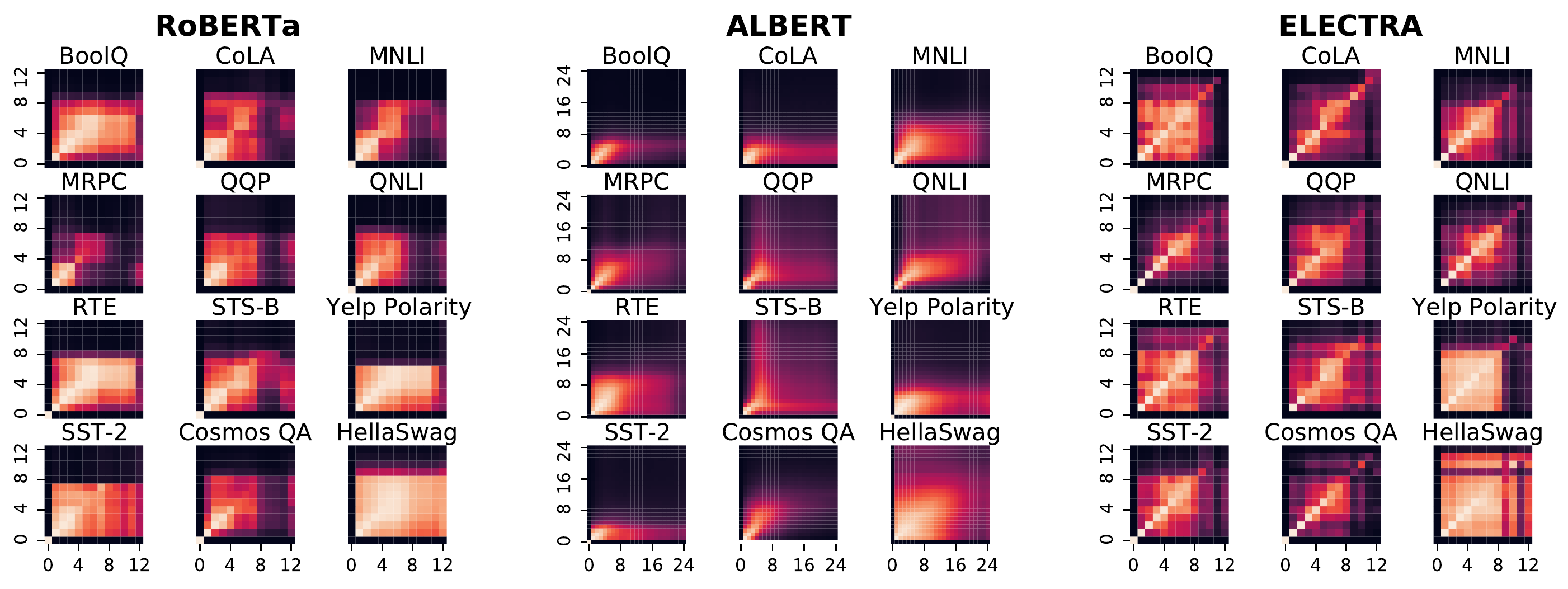}
    \caption{
        CKA representation similarity for FT--ORIG.
        Task-tuned layers are on the Y-axis, untuned layers in the X-axis.
        \CLS representations of the top few layers RoBERTa and ALBERT models are highly dissimilar to those of the pretrained model at any layer.
    }
    \label{fig:cka-taskbase-cls}
\end{figure*}

Figure~\ref{fig:cka-taskbase-cls} shows the FT--ORIG plots for all tasks and models.

\begin{figure*}[t]
    \centering
    \includegraphics[width=\textwidth]{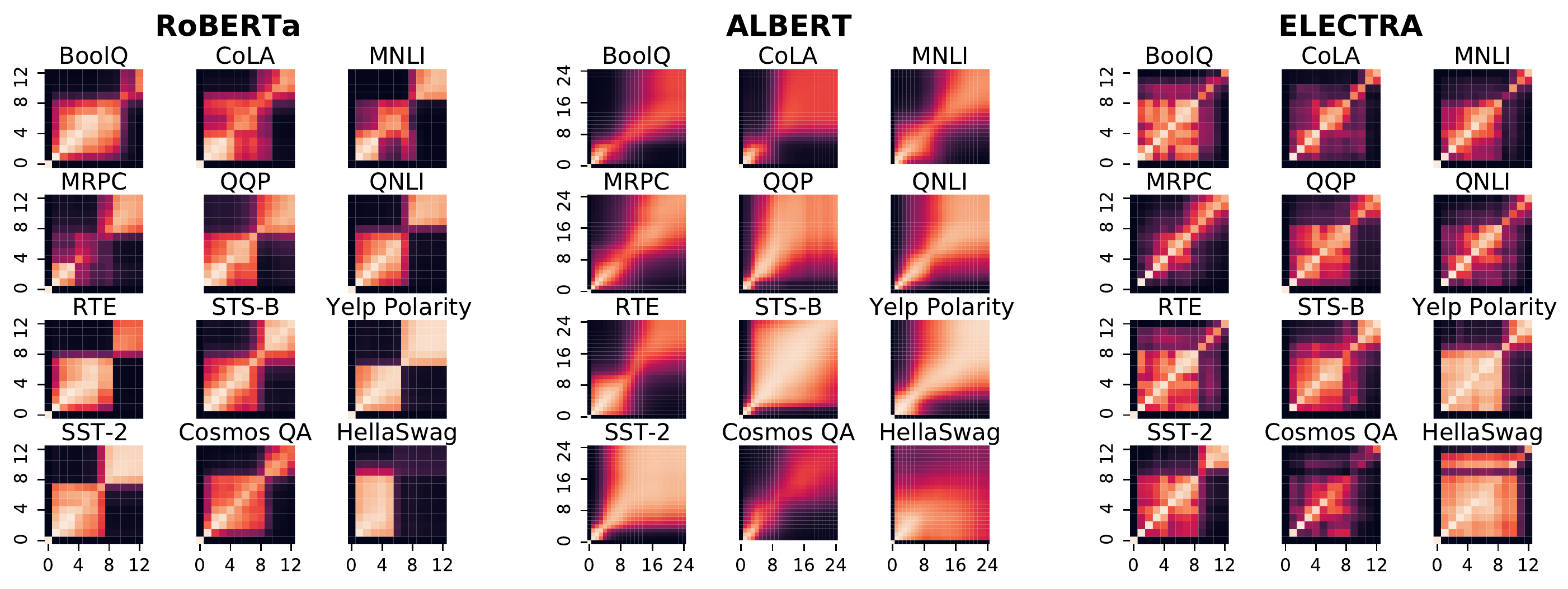}
    \caption{
        CKA representation similarity for FT[1]--FT[2].
        RoBERTa and ALBERT task models exhibit a `block diagonal` structure to representation similarity of \CLS tokens, indicating in particular that the representations of the top few layers are highly similar.
        Plots for tasks that do not use the \CLS token are dimmed.
    }
    \label{fig:cka-tasktask2-cls}
\end{figure*}

Figure~\ref{fig:cka-tasktask2-cls} shows the FT[1]--FT[2] plots for all tasks and models.

\begin{figure*}[t]
    \centering
    \includegraphics[width=\textwidth]{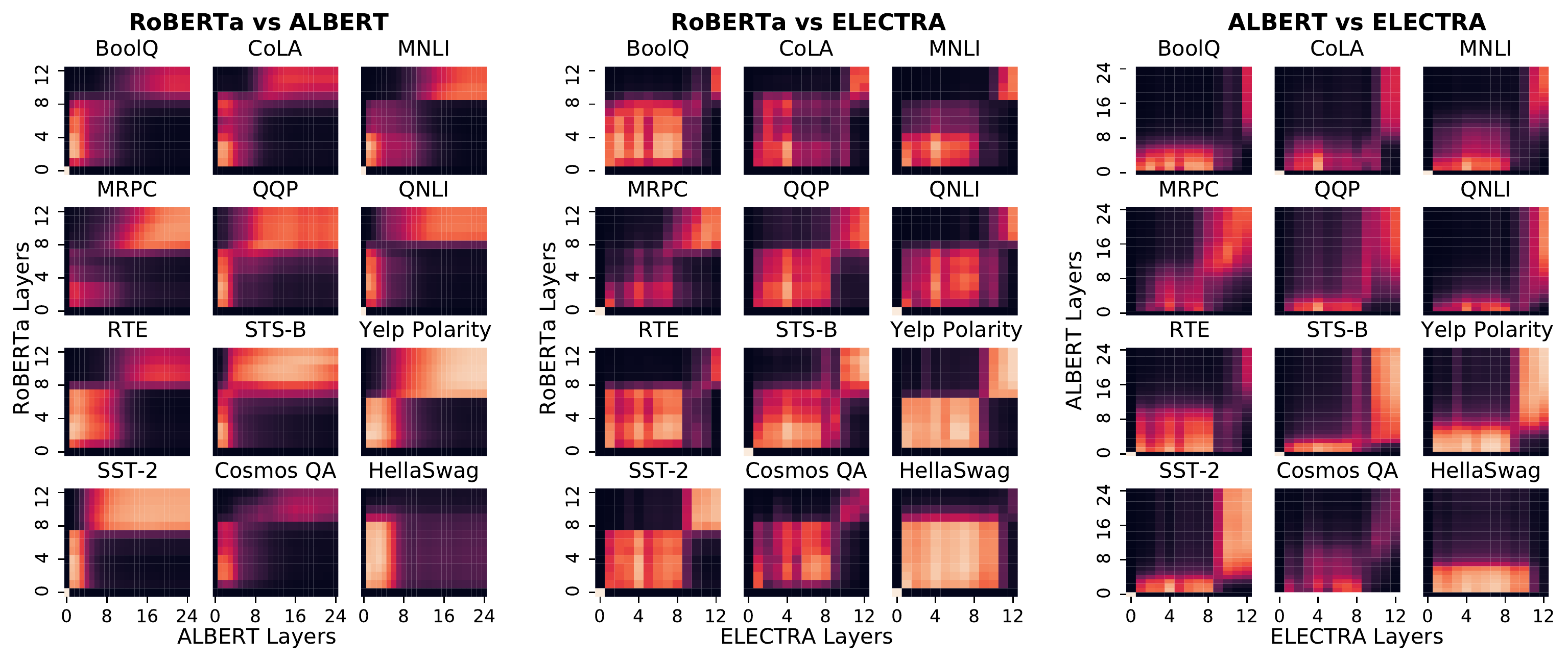}
    \caption{
        CKA representation similarity comparing \CLS representations cross models. The upper right blocks indicate the representations in the earlier and the later layers are similar even across models.
    }
    \label{fig:cka-cross}
\end{figure*}

Figure~\ref{fig:cka-cross} computes representation similarity \textit{between} models.

\end{document}